\documentclass{elsart}
\usepackage{algorithm,algorithmic,amsmath,amssymb,framed,graphicx,multirow}
\usepackage[switch,pagewise]{lineno}
\usepackage[usenames]{color}
\usepackage[ pdfstartview=FitH, CJKbookmarks=true,  bookmarksnumbered=true, bookmarksopen=true, pdfborder=001,  linkcolor=black, anchorcolor=black,  citecolor=black ]{hyperref}

\renewcommand{\algorithmiccomment}[1]{// #1}

\newcommand{\tabincell}[2]{\begin{tabular}{@{}#1@{}}#2\end{tabular}}
\begin{document}
\begin{frontmatter}

\title{Effective reinforcement learning based local search for the maximum $k$-plex problem}

\author[wuhan]{Yan~Jin},
\ead{jinyan@mail.hust.edu.cn}
\author[UK1]{John~H.~Drake},
\author[UK2]{Una~Benlic},
\author[wuhan]{Kun~He}
\address[wuhan]{School of Computer Science, Huazhong University of Science \& Technology, Wuhan,China}
\address[UK1]{Operational Research Group, Queen Mary University of London, London, United Kingdom}
\address[UK2]{SATALIA, 40 High Street, Islington High Street, London, United Kingdom}

\maketitle

\begin{abstract}
The maximum $k$-plex problem is a computationally complex problem, which emerged from graph-theoretic social network studies. This paper presents an effective hybrid local search for solving the maximum $k$-plex problem that combines the recently proposed breakout local search algorithm with a reinforcement learning strategy. The proposed approach includes distinguishing features such as: a unified neighborhood search based on the swapping operator, a distance-and-quality reward for actions and a new parameter control mechanism based on reinforcement learning. Extensive experiments for the maximum $k$-plex problem ($k = 2, 3, 4, 5$) on 80 benchmark instances from the second DIMACS Challenge demonstrate that the proposed approach can match the best-known results from the literature in all but four problem instances. In addition, the proposed algorithm is able to find 32 new best solutions.

\end{abstract}

\end{frontmatter}

\section{Introduction}
\label{Sec_Intro}
Let $G = (V, E)$ be a simple undirected graph, where $V = \{v_1, \ldots, v_n\}$ is the set of vertices and $E \subset V \times V$ is the set of edges. A $k$-plex for a given positive integer $k$ is a subset $S$ of $V$ such that each vertex of $S$ is adjacent to at least $|S| - k$ vertices in the subgraph $G[S] = (S, E \cap S \times S)$ induced by $S$. Formally, let $N(v)$ be the set of adjacent vertices $v_i$ of $v$ (i.e., $\{v_i, v\} \in E$), the maximum $k$-plex problem with any fixed $k \geq 1$ ($k \in Z^+$) aims to find a $k$-plex $S$ of maximum cardinality, such that $\forall v \in S, |N(v) \cap S| \geq |S| - k$. The $k$-plex problem first aroses in the context of graph theoretic social network \cite{seidman1978graph} and has become popular in several other contexts \cite{berry2004emergent,boginski2014network,gibson2005discovering}. The $k$-plex problem with any fixed positive integer $k$ is an NP-complete problem \cite{balasundaram2011clique}, it reduces to the well-known maximum clique problem (MC) when $k = 1$, one of Karp's 21 NP-complete problems \cite{karp1972reducibility}. The $k$-plex problem has a number of applications in information retrieval, code theory, signal transmission, social networks, classification theory amongst others \cite{du2007community,krebs2002mapping,newman2001structure,xiao2017generalization}.

Due in part to the wide variety of real-world applications, increased research effort is being devoted to solving this problem. Over the past few years, several exact algorithms have been proposed for finding the maximum $k$-plex of a given graph \cite{balasundaram2011clique,mcclosky2012combinatorial,moser2012exact,trukhanov2013algorithms,xiao2017generalization}. These methods can find optimal solutions for graphs with around a thousand vertices in a reasonable amount of computing time (within 3 hours). However, they often fail to solve larger instances of the problem. On the other hand, several heuristic approaches have also been presented, which are able to find high-quality solutions for larger problem instances \cite{gujjula2014hybrid,miao2012cluster,zhou2017frequency}. Two of these approaches \cite{gujjula2014hybrid,miao2012cluster} are based on the general GRASP framework \cite{resende2010greedy}, whilst the other \cite{zhou2017frequency} uses the tabu search metaheuristic.

One notices that, compared to the research effort on the MC problem, studies for the maximum $k$-plex problem are more recent and less abundant. In this work, we are interested in approximately solving the representative large scale $k$-plex instances by presenting an effective heuristic approach. Moreover, the reinforcement learning techniques are shown to be able to improve the performance of local search algorithms \cite{benlic2017hybrid,zhou2016reinforcement}, hereby, we are also interested in investigating the reinforcement learning based local search for solving the maximum $k$-plex problem. We apply the recent breakout local search and the reinforcement learning together to explore the search space (denoted as BLS-RLE). It uses descent search to discover local optima before applying sophisticated diversification strategies to move to unexplored regions of the search space. BLS-RLE integrates several distinguishing features to ensure that the search process is effective. Firstly, a reinforcement learning technique is applied to adaptively and interdependently control three parameters deciding the type of perturbation to be applied and the magnitude of that perturbation. Secondly, the search is driven by a unified ($q$,1)-swap($q$ $\in$ $Z$) operator, used to explore the constrained neighborhoods of the problem. Finally, a distance-and-quality reward is used to maintain a high-quality set of parameters based on previous experience.

We evaluate BLS-RLE for the $k$-plex problem (with $k = 2, 3,4, 5$) on a set of 80 large and dense graphs from the second DIMACS Challenge benchmark. Comparisons are performed to a number of state-of-the art methods from the literature, and the computational results show that BLS-RLE is able to achieve the best-known results for all the instances tested except 4 cases. In particular, BLS-RLE finds 11 new best solutions for $k = 2$, 7 for $k = 3$, 7 for $k = 4$ and 7 for $k = 5$ respectively.

The rest of the paper is organized as follows. Section \ref{Sec_BLSRLE} presents the proposed reinforcement learning based local search algorithm for solving the maximum $k$-plex problem. Section \ref{Sec_ExperimentalSetup} shows the computational results of BLS-RLE for the $k$-plex problem with $k = 2, 3,4, 5$ and comparisons with the state-of-the-art algorithms in the literature. Before concluding, Section \ref{SecAnalysis} investigates and analyzes the influence of reinforcement learning for the local search algorithm.

\section{The reinforcement learning based local search algorithm}
\label{Sec_BLSRLE}
The proposed reinforcement learning based local search algorithm (BLS-RLE) follows the recent general learning based local search framework, which was first introduced by \cite{benlic2017hybrid} and applied to the Vertex Separator Problem. BLS-RLE combines an intensification stage that applies descent local search, with a distinctive diversification stage that adaptively selects between two or more perturbation operators with a particular perturbation magnitude. The type of perturbation operator (denoted as $e$), the number of perturbation moves (the depth of perturbation, denoted as $l$) and the degree of random perturabtion (denoted as $b$) are three important self-adaptive parameters that control the degree of diversification during perturbation. The combination of these three parameters by determining the values of $l$, $e$ and $b$ independently may not constitute the most suitable degree of diversification required at one stage of the search. And \cite{Hoos2011Automated} highlights the automated configuration and parameter tuning techniques are very important for effectively solving difficult computational problems. Hence, unlike the parameter tuning techniques introduced in \cite{Hoos2011Automated}, BLS-RLE uses a parameter control mechanism based on reinforcement learning \cite{auer2002finite} to adaptively and interdependently determine the value of parameters.

\subsection{General procedure}
\label{Subsec_GeneralProcedure}
The overall BLS-RLE algorithm is summarized in Algorithm \ref{Algo_BLSRLE}. Following the Prelearning phase, the first step of each iteration of BLS-RLE consists of selecting an action (parameter triple $(l,e, b)$) according to the Softmax action-selection rule \cite{sutton1998introduction} (line 4). Although we use Softmax, any action selection model could be used as an alternative. The diversification procedure is then applied to the current local optimum $S$ using $(l, e, b)$ (line 5), followed by the $Intensification$ (descent local search) to improve the quality of the perturbed solution $S'$ (line 6). Furthermore, a global variable $S^*$ is used to record the best $k$-plex solution discovered during the search (lines 7-9). Finally, BLS-RLE applies the $DetermineParameterTripleReward$ procedure to determine the reward $r$ for the selected action (parameter triple) with regard to $S$ by considering both the quality and the diversity criterion (line 10). Finally, the action value corresponding to $(l,e, b)$ is updated with $r$ (line 11).

\begin{algorithm}[h]
\caption{General procedure of BLS-RLE for the maximum $k$-plex problem}\label{Algo_BLSRLE}
\begin{algorithmic}[1]
   \REQUIRE  A graph $G = (V, E)$, an integer $k$
   \ENSURE The largest $k$-plex $S^*$ found
   \STATE $(P_{learn}, S)$ $\gets$ $Prelearning$();  /* $P_{learn}$ is a set of parameter triples $\{(l_0,e_0,b_0), (l_1,e_1,b_1), \ldots\}$, $S$ is a feasible $k$-plex */
   \STATE $Max \gets |S|;$
   \STATE $Iter \gets 0$;
   \REPEAT
   \STATE $(l,e,b) \gets SelectParameterTriple()$;
   \STATE $S' \gets Diversification(l,e,b,S)$;
   \STATE $S \gets Intensification(S')$;
   \IF{$|S| > |S^*|$}
      \STATE $S^* \gets S; Max \gets |S^*|;$
   \ENDIF
   \STATE $r \gets DetermineParameterTripleReward(S)$;
   \STATE $ApplyReward(r)$;
   \IF{$Iter > \epsilon$}
      \STATE $UpdateParameterTriple()$;
      \STATE $Iter \gets 0$;
   \ENDIF
   \STATE $Iter \gets Iter +1$;
   \UNTIL{a stop criterion is met}
   \RETURN the best found $S^*$;
\end{algorithmic}
\end{algorithm}

A parameter triple $(l,e,b)$ determines the perturbation type, perturbation magnitude and degree for diversification, where $l$ represents the number of perturbation moves, $e$ represents the probability of selecting one type of perturbation operator over another and $b$ represents the degree of random perturbation to adjust the strength of random peturbation. The set $P_{all}$ contains all possible parameter triples, i.e., $P_{all} = \{(l_0,e_0,b_0),\ldots,(l_m,e_m,b_m)\}$ where $m$ is the total number of triples to be generated. Here we use one of two types of perturbation operator, \emph{directed} or \emph{random}. A novel method for generating parameter triples is proposed, with the number of perturbation moves $l in \{l_i\}$ takes the following piecewise function (Eq. (\ref{eqPiecewise})). The linear nature of the first component reflects a more fine-grained approach to diversification, using a larger number of values when the perturbation level is low. As the perturbation level increases, the number of potential values for $l$ will decrease, with the values that they take increasing exponentially. This provides a greater level of diversification within a smaller number of potential $l$ values at the higher end.
\begin{equation} \small \label{eqPiecewise}
l_i=\left\{
\begin{array}{rcl}
i+2 & & {0 \leq i < 30}\\
2^{i-25} & & {30 \leq i < 33}
\end{array} \right.
\end{equation}

%When the search is stagnated, the small number of perturbation moves are quite useful, either one maybe strong enough to make the search escape from local optima. But if the perturbation moves are large (in our case, we set the perturbation moves are larger than 30), we do not need to try them one by one, we could use several sampling number to have a try

This method for defining the set of potential $l$ values and subsequently triples in $P_{all}$ is in contrast to previous work \cite{benlic2017hybrid}, where a linear relationship between consecutive $l$ values is maintained throughout the range. As for the parameters $e$ and $b$, the value of $e$ ranges from 95 to 100, and $b$ ranges 70 to 90 respectively.

The algorithm uses a one-time $Prelearning$ procedure to evaluate the degree of diversification introduced by each parameter triple $(l, e, b) \in P_{all}$ that can be applied throughout the search process. For this purpose, the degree of diversification refers to the ability of the search to discover new local optima. To limit the number of possible actions and to reduce the time required to learn, BLS-RLE maintains a small subset $P_{learn} \subset P_{all}$ of $\kappa$ parameter triple, representing potential actions that can be performed at that time. $P_{learn}$ is periodically updated (every $\epsilon$ iterations) with a new parameter triple from $P_{all} \setminus P_{learn}$ based on the action values learned during the Prelearning procedure.

\subsection{Prelearning procedure}
\label{Subsec_PrelearningProcedure}
Iterated Local Search is used to assess the diversification capability of each parameter triple $(l,e,b)$, based on how frequently a new local optimum is found. The number of times a previously encountered local optima is visited by each triple is recorded, with the value for each triple initially set to 0. The detailed procedure is as follows:
\begin{enumerate}
\item A perturbation operator is selected and applied, based on the parameter $e$. %If a random probability is smaller than $e$, directed perturbation is applied, favoring moves that minimize the degradation of the cardinality of the given $k$-plex. Otherwise random perturbation is used, selecting a random solution that is better than a given acceptance condition.
%Irrespective of the type of perturbation operator chosen,
The perturbation operator makes $l$ moves, with a hash table used to record recently visited local optima as historical information.
\item Following this, descent local search attempts to improve the solution, returning the local optimum.
\item Finally, a check is performed to see if the local optimum has already been encountered. If so, the count of revisited local optima for the corresponding triple is increased by 1.
\item Repeat steps 1 to 3 until each parameter triple $(l,e,b)$ has been used.
\end{enumerate}
This process is repeated $\alpha$ times, where $\alpha$ is chosen to provide enough samples from which a ranking of triples can be derived. Once completed, the parameter triples in $P_{all}$ are sorted in ascending order of the number of times they revisited previously encountered local optima, with the $n$ best-ranked triples kept in a set $P_{learn}$. Note that here $n$ is experimentally set to 2375, and that this ranking will also be used for the reward and value function.

\subsection{Intensification of search}
\label{Subsec_IntensificationOfSearch}
The intensification stage of BLS-RLE aims to find better solutions by descent local search. For this purpose, BLS-RLE employs the (0,1)-swap move operator to try to improve a solution. Recall that $N(v)$ is the set of adjacent vertices of $v$, let $V \backslash S$ be the complementary set of $S$ and $N(S)$ be the set of vertices in $V \backslash S$ that are adjacent to at least one vertex in $S$, i.e., $N(S) = \cup_{v \in S} N(v)\backslash S$. For a vertex $v$ in $S$, if $|N(v) \cap S| = |S|-k$, then $v$ is called a \emph{critical} vertex. The set $NS_0$, shown in Eq. (\ref{eq1}), consists of all vertices in $N(S)$ that are connected to at least $|S|-k+1$ vertices in $S$ and are also connected to all of the \emph{critical} vertices $C$ \cite{trukhanov2013algorithms,zhou2017frequency}. %i.e., $NS_0 = \{v \in N(S): |N(v) \cap S| \geq |S| - k + 1, C \backslash N(v) = \emptyset \}$.
Descent local search includes a new vertex from $NS_0$ to increase the cardinality of the current $S$, while maintaining the feasibility of the $k$-plex. When $NS_0$ is empty, the intensification stage is complete and the local optimum found is returned.
\begin{equation} \small \label{eq1}
NS_0 = \{v \in N(S): |N(v) \cap S| \geq |S| - k + 1, C \backslash N(v) = \emptyset \}
\end{equation}

\subsection{Perturbation operators}
\label{Subsec_PerturbationOperators}
In order to escape from local optima, %a dedicated diversification procedure applies
directed or random perturbations are used to guide the search towards unexplored regions. Directed perturbation aims to minimize the degradation of the $k$-plex cardinality, while random perturbation aims to move the search away from the current location. The choice of operator is based on the probability $e$ of using a directed perturbation operator, with random perturbation applied otherwise (with the corresponding probability 1 - $e$). If the random perturbation is chosen, the value of parameter $b$ controls the strength of random perturbation (the random perturbation is compared to a random start when $b \approx 0$). The number of perturbation moves is defined as $l$, with $l$, $e$ and $b$ controlled interdependently by the reinforcement learning strategy.

The proposed approach uses a unified $(q,1)-swap$ ($q = 0, 1, 2, \ldots$) move to explore the search space. Four sets $NS_i$ $(i=0, 1, 2, 3)$ are involved in this procedure as given below, $NS_0$ is already presented above. The $NS_1$ set consists of vertices $v$ that are adjacent to at least $|S|-k$ vertices in $S$, where $u$ is the unique \emph{critical} vertex not adjacent to $v$, as shown in Eq. (\ref{eq2}). $NS_2$ consists of vertices $v$ that are adjacent to exactly $|S| - k$ vertices in $S$. These $|S| - k$ vertices should include all of the \emph{critical} vertices,  as shown in Eq. (\ref{eq3}) \cite{zhou2017frequency}. A corresponding exchanged vertex $u$ is randomly selected from $S \backslash N(v)$. The vertex from $N_1 \cup N_2$ could be employed for the (1,1)-swap move operator, such that the search is guided to a new region while leaving the quality of solution unchanged. The $NS_{>2}$ consists of the vertex $v$, such that $v$ is in the set of $V \backslash S$ and $v$ is not included in the sets $NS_0$, $NS_1$ and $NS_2$. A corresponding exchanged vertex $u$ is selected from $S$ that will cause the $k$-plex to be infeasible after including $v$. The vertex from $NS_{3}$ could be used for the $(q,1)$-swap ($q \geq 2$) move in order to discover new and promising regions of the search space.
\begin{equation}\small \label{eq2}
NS_1 = \{v \in N(S): |N(v) \cap S| \geq |S| - k, |C \backslash N(v)| = 1 \}
\end{equation}
\begin{equation} \small \label{eq3}
NS_2 = \{v \in N(S): |N(v) \cap S| = |S| - k, C \backslash N(v) = \emptyset \}
\end{equation}
\begin{equation} \small \label{eq4}
NS_{3} = \{v \in V \backslash ({S \cup NS_0 \cup NS_1 \cup NS_2}) \}
\end{equation}

Directed perturbation %within the diversification procedure
applies a move from $NS_0 \cup NS_1 \cup NS_2 \cup NS_3$, favoring moves that minimize the degradation of the cardinality of the current $k$-plex. Random perturbation applies a random move from $NS_0 \cup NS_3$. Each move is only accepted when the quality is not worse than a given threshold, determined by the value of parameter $b$. A tabu list is used to prevent the search returning to previously visited locations. More precisely, each time a vertex is removed from the $k$-plex when employing the ($q$,1)-swap ($q \geq 2$), this vertex is prevented from moving back to $S$ for the next $tt_1$ iterations ($tt_1$ is called tabu tenure and $tt_1$ = 7). On the other hand, each time a vertex is removed from the $k$-plex when employing the (1,1)-swap, the tabu tenure is set to $tt_2 = 7 + random(|NS_1 \cup NS_2|)$ where $random(|NS_1 \cup NS_2|)$ is a random integer from 1 to $|NS_1 \cup NS_2|$.

\subsection{Reward and value functions}
\label{Subsec_RewardAndValueFunctions}
After a locally optimal solution $S$ is returned by descent local search, %and a diversification stage with the parameter triple $(l,e)$,
we apply a distance-and-quality reward for each parameter triple (action). This reward considers both the quality of solution and the degree to which new areas of the search space are explored. The distance-and-quality reward $r_i$ is given in Eq. (\ref{eq5}) \cite{benlic2013breakout}. The core motivation behind this strategy is to give the highest reward when new local optima are discovered with the minimum amount of diversification introduced. Recall that the parameter triples in $P_{all}$ are sorted in increasing order in the $Prelearning$ phase, the distance is determined by $\kappa - i$ where $i$ is the index $i$ of the parameter triple $(l,e,b)_i$ when $S$ is not stored in the hash table. The quality is evaluated by $ (1-\frac{|S|-|S^*|}{|S^*|})^2 \times 10$ where $|S|$ is the cardinality of the current $k$-plex $S$ and $S^*$ is the best solution found so far during the search.

\begin{equation} \small \label{eq5}
r_i =
  \left\{
   \begin{array}{c}
   0,  \qquad \qquad  \qquad  \qquad   \textrm{If $S$ is already in the hash table}\\
   \delta_1 \times (\kappa - i) + \delta_2 \times (1-\frac{|S|-|S^*|}{|S^*|})^2 \times 10 , \textrm{Otherwise}  \\
   \end{array}
  \right.
  \end{equation}

As soon as the reward $r_i$ for the parameter triple $(l,e,b)_i$ is computed, the $ApplyReward$ procedure is used to update the credit attributed to the action $(l,e,b)_i$ and accordingly compute the action value $\omega_i$. The values of $\omega_i$ are initialized to 1, and the 100 latest rewards ascribed to $(l,e,b)_i$ are used to estimate the credit that quantifies the performance of a particular action in a given period. According to this performance summary, the action value is updated.

\subsection{Update of the parameter triple learning list}
\label{Subsec_UpdateParameterTriple}
In order to prevent the search process from premature convergence, the parameter triple learning list should be updated periodically. Afterwards, the Softmax action-selection model \cite{sutton1998introduction} is used to select an action for the next iteration of the search.

\begin{figure}[!h]\centering
\includegraphics[scale=0.5]{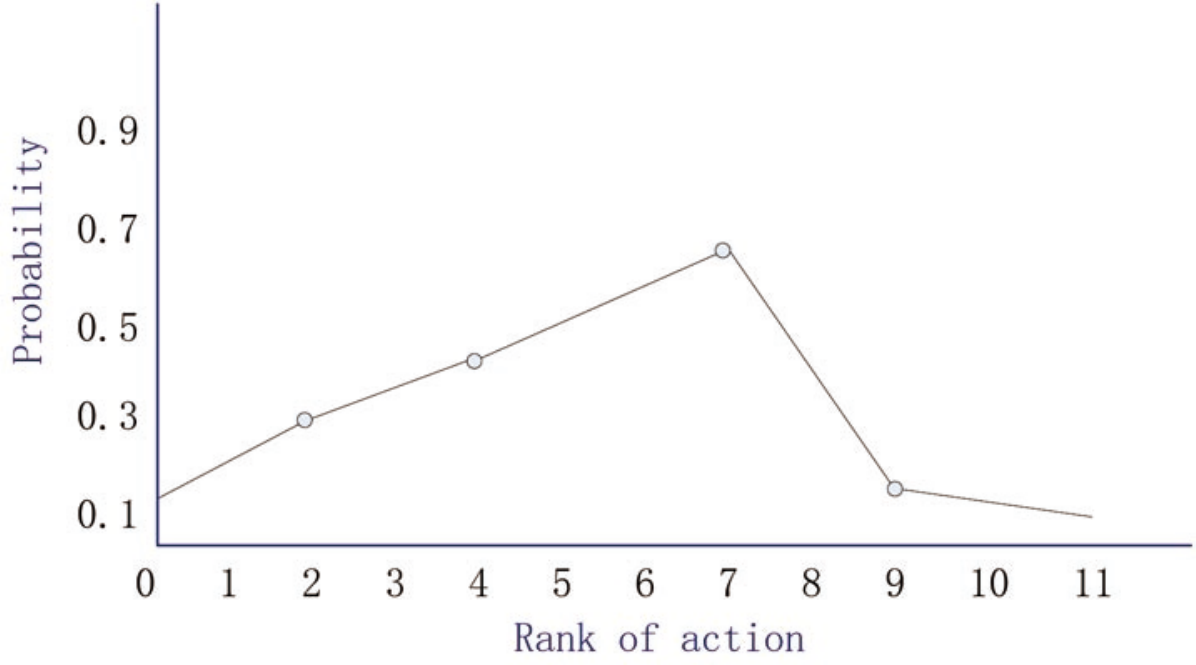}
\caption{An illustrative example for estimating probability.}\label{fig_Example}
\end{figure}

The Softmax-based approach applies the Gibbs or Boltzmann distributions to assign a probability to each action in $P_{learn}$ \cite{benlic2013breakout}. The updating rule for the triple learning list first selects the worst parameter triple $a_w \in P_{learn}$, which has the lowest probability of selection. Then, the updating rule determines one promising parameter triple in $P_{all} \backslash P_{learn}$. According to the probabilities of actions in $P_{learn}$, we estimate the probability of each action in $P_{all} \backslash P_{learn}$ by a linear correspondence. For example, given 12 actions in $P_{all}$, 4 actions are selected from $P_{learn}$ and the probability of these 4 actions are known. We use a line to connect two adjacent actions whose probabilities are known, as shown in Figure \ref{fig_Example}. For action $a_0 \in P_{all} \backslash P_{learn}$, the estimated probability of $a_0$ is obtained by the probability of the first action $a_0'$ in $P_{learn}$ divided by the index of action $a_0'$ in $P_{all}$. Similarly, the estimated probability of $a_{11} \in P_{all} \backslash P_{learn}$ is obtained by the probability of the last action $a_3'$ in $P_{learn}$ divided by the difference between 11 and the index of action $a_3'$ in $P_{all}$. Finally, we estimate the probability of all vertices in $P_{all} \backslash P_{learn}$ and select a vertex with a high probability to replace the worst action $a_w$ in $P_{learning}$. After the parameter triple learning list is updated, the actions in $P_{learn}$ are re-sorted, the action values are reset to 1, and the next round of search is triggered.

\section{Experimental Set-up}
\label{Sec_ExperimentalSetup}

In order to evaluate the performance of %the proposed
BLS-RLE, we conduct experiments on 80 benchmark instances\footnote{https://turing.cs.hbg.psu.edu/txn131/clique.html} from the Second DIMACS Implementation Challenge \cite{johnson1996cliques}. These instances were first established for the maximum clique (MC) problem and are frequently used for evaluating MC algorithms. As previously mentioned, the maximum $k$-plex problem can be reduced to the MC, hence these instances are also quite popular and challenging for evaluating solution methods to this problem.

The BLS-RLE algorithm is implemented in C++ and complied using g++ with the `-O3' option under GNU/Linux running on an Intel Xeon E5440 processor (2.83GHz and 4GB RAM). When the DIMACS machine benchmarks\footnote{ftp://dimacs.rutgers.edu/pub/dsj/clique/} are run on our machine, the run time required is 0.44, 2.63 and 9.85 seconds for graphs r300.5, r400.5 and r500.5 respectively. Each instance is run 20 times independently. The algorithm stops when a fixed cutoff time (180 seconds) is met. This experimental protocol is also used by a state-of-the-art heuristic in the literature \cite{zhou2017frequency}. BLS-RLE uses the self-adaptive parameters given in Table \ref{table_parameterSettings}, all computational results were obtained with the same parameters. The adopted parameter settings are inspired by a previous method from the literature \cite{benlic2017hybrid}, where a parameter sensitivity analysis during the search guidance was provided.

\begin{table}[h] \centering
\begin{scriptsize}
\caption{Parameter settings of BLS-RLE}
\label{table_parameterSettings}
%\begin{tabular}{lcrrr}
\begin{tabular}{p{0.9cm}p{5.8cm}p{0.4cm}}
\hline
Parameter & Description  & Value   \\
\hline
$\kappa$	&the size of the learning parameter triple set $P_{learn}$	&	6		\\
$\alpha$	&the number of pre-learning iterations for each triple in $P_{all}$	&	100		\\
$\tau$	&the temperature of the Softmax-based adaptive procedure	&	2		\\
$\epsilon$	&the update frequency of the parameter triple learning list	&	4000		\\
$\delta_1$	&coefficient for reward function	&	2	\\
$\delta_2$	&coefficient for reward function	&	1	\\
%$tt_1$	&coefficient for tabu tenure  	&	7	\\
%$tt_2$	&coefficient for tabu tenure  	&	7+$random(|NC_2|)$	\\
%$\phi$	&coefficient used by the strategy that updates $P_{learn}$  	&	2	\\
%$\omega$	&the number of latest rewards considered for computing the empirical quality estimate  	&	100	\\
\hline
\end{tabular}
\end{scriptsize}
\end{table}

In addition to BLS-RLE we also provide the results of BLS-RND, a variant of BLS where the values of the two parameters, $l$, $e$ and $b$ are determined randomly. \cite{karafotias2013parameter} highlighted the importance of comparing adaptive parameter control mechanisms with random sampling of the parameter space. When introducing a dynamic parameter control strategy, rather than using static parameter values, it is not clear whether it is the adaptive strategy or simply that the value of parameter is changing dynamically over time that is having an effect on performance.

\section{Experimental Results}
\label{Sec_ExperimentalResults}

To assess the performance of BLS-RLE, we compare it with some state-of-the-art algorithms from the literature \cite{balasundaram2011clique,mcclosky2012combinatorial,moser2012exact,trukhanov2013algorithms,zhou2017frequency}. The experimental platform of \cite{balasundaram2011clique} was performed on Dell Precision PWS690 machine with a 2.66GHz Xeon Processor, 3 GB RAM and 120 GB HDD. The experimental platform of \cite{mcclosky2012combinatorial,moser2012exact} was performed on a 2.2 GHz Dual-Core AMD Opteron Processor with 3GB memory, a AMD Athlon 64 3700+ machine with 2.2 GHz and 3GB memory respectively. The experimental platform of \cite{trukhanov2013algorithms} was run on an Intel $Core^{TM}$2 Quad 3 GHz Processor with 4GB RAM, a 2.5 GHz Intel Core i5-3210M processor with 4GB memory respectively. The experimental platform of \cite{zhou2017frequency} was run on an AMD Opteron 4184 with 2.8 GHz Processor with 2GB RAM.
Table \ref{tabledet2}, \ref{tabledet3}, \ref{tabledet4} and \ref{tabledet5} compare BLS-RLE to several recent best-performing algorithms and the heuristic method proposed in \cite{zhou2017frequency} (FD-TS) for $k = 2, 3, 4, 5$, covering the best known results for all of the instances tested.
%The full results for $k$ = 3, 4, 5 have been omitted here due to space limitations, however they are available online at: [ADD WEBLINK IF PAPER IS ACCEPTED].

\renewcommand{\baselinestretch}{0.5}\large\normalsize
\begin{table}[htp]
\begin{center}
\begin{scriptsize}
\caption{Comparisons of BLS-RLE for the $k$-plex when $k$=2}
\label{tabledet2}
%\begin{tabular}{p{1.1cm}p{0.4cm}p{0.4cm}p{1.1cm}p{0.4cm}p{1.1cm}p{0.3cm}p{0.0cm}}
\begin{tabular}{p{1.2cm}rlrp{1mm}lr}
\hline
\tabincell{c}{Instance} &BEV    &\multicolumn{2}{c}{FD-TS} &&\multicolumn{2}{c}{BLS-RLE} \\						
\cline{3-4} \cline{6-7}							
&  &   Max (Avg.) & Time(s)    & & Max (Avg.) & Time(s)   \\						
\hline															
\tabincell{c}{	brock200\_1	}&	25			&	26	&	0.15	&	&	26	&	0.04	\\
\tabincell{c}{	brock200\_2	}&	13			&	13	&	0.01	&	&	13	&	0.01	\\
\tabincell{c}{	brock200\_3	}&	17			&	17	&	0.02	&	&	17	&	0.02	\\
\tabincell{c}{	brock200\_4	}&	20			&	20	&	0.06	&	&	20	&	0.03	\\
\tabincell{c}{	brock400\_1	}&	23			&	30	&	0.42	&	&	30	&	0.16	\\
\tabincell{c}{	brock400\_2	}&	27			&	30	&	0.51	&	&	30	&	0.11	\\
\tabincell{c}{	brock400\_3	}&	-			&	30	&	0.35	&	&	30	&	0.20	\\
\tabincell{c}{	brock400\_4	}&	27			&	33 (31.2)	&	64.18	&	&	\emph{31}	&	23.83	\\
\tabincell{c}{	brock800\_1	}&	-			&	25	&	10.90	&	&	25	&	2.26	\\
\tabincell{c}{	brock800\_2	}&	-			&	25	&	11.36	&	&	25	&	1.82	\\
\tabincell{c}{	brock800\_3	}&	-			&	25	&	12.68	&	&	25	&	2.28	\\
\tabincell{c}{	brock800\_4	}&	-			&	26 (25.55)	&	56.21	&	&	26 (25.9)	&	51.07	\\
\tabincell{c}{	C1000.9	}&	-			&	81 (80.55)	&	39.65	&	&	\textbf{82 (81.75)}	&	51.66	\\
\tabincell{c}{	C125.9	}&	-			&	43	&	0.00	&	&	43	&	0.01	\\
\tabincell{c}{	C2000.5	}&	-			&	19 (18.95)	&	27.96	&	&	\textbf{20 (19.05)}	&	15.00	\\
\tabincell{c}{	C2000.9	}&	-			&	90 (88.9)	&	65.21	&	&	\textbf{93 (91.7)}	&	76.41	\\
\tabincell{c}{	C250.9	}&	-			&	55	&	8.43	&	&	55	&	12.96	\\
\tabincell{c}{	C4000.5	}&	-			&	20	&	39.23	&	&	\textbf{21 (20.15)}	&	11.86	\\
\tabincell{c}{	C500.9	}&	-			&	69	&	10.67	&	&	69	&	2.78	\\
\tabincell{c}{	c-fat200-1	}&	12			&	12	&	0.00	&	&	12 (10.75)	&	0.01	\\
\tabincell{c}{	c-fat200-2	}&	24			&	24	&	0.01	&	&	24 (22.25)	&	0.02	\\
\tabincell{c}{	c-fat200-5	}&	58			&	58	&	0.01	&	&	58 (57.4)	&	0.01	\\
\tabincell{c}{	c-fat500-10	}&	126			&	126	&	0.08	&	&	126 (124.95)	&	0.02	\\
\tabincell{c}{	c-fat500-1	}&	14			&	14	&	0.00	&	&	14 (12.4)	&	0.01	\\
\tabincell{c}{	c-fat500-2	}&	26			&	26	&	0.00	&	&	26 (25.2)	&	0.01	\\
\tabincell{c}{	c-fat500-5	}&	64			&	64	&	0.02	&	&	64 (62.45)	&	0.01	\\
\tabincell{c}{	DSJC1000\_5	}&	-			&	18	&	26.61	&	&	18	&	13.37	\\
\tabincell{c}{	DSJC500\_5	}&	-			&	16	&	0.29	&	&	16	&	0.18	\\
\tabincell{c}{	gen200\_p0.9\_44	}&	-			&	53	&	0.14	&	&	53	&	0.38	\\
\tabincell{c}{	gen200\_p0.9\_55	}&	-			&	57	&	0.02	&	&	57	&	0.04	\\
\tabincell{c}{	gen400\_p0.9\_55	}&	-			&	68 (67.7)	&	65.76	&	&	68	&	40.14	\\
\tabincell{c}{	gen400\_p0.9\_65	}&	-			&	73 (71.6)	&	28.01	&	&	\textbf{74 (72.75)}	&	44.32	\\
\tabincell{c}{	gen400\_p0.9\_75	}&	-			&	79 (78.05)	&	30.62	&	&	\textbf{80 (79.1)}	&	35.22	\\
\tabincell{c}{	hamming10-2	}&	512			&	512	&	8.97	&	&	512	&	1.96	\\
\tabincell{c}{	hamming10-4	}&	41			&	48	&	1.53	&	&	48	&	0.45	\\
\tabincell{c}{	hamming6-2	}&	32			&	32	&	0.00	&	&	32	&	0.01	\\
\tabincell{c}{	hamming6-4	}&	6			&	6	&	0.00	&	&	6	&	0.01	\\
\tabincell{c}{	hamming8-2	}&	128			&	128	&	0.09	&	&	128	&	0.02	\\
\tabincell{c}{	hamming8-4	}&	16			&	16	&	0.01	&	&	16	&	0.01	\\
\tabincell{c}{	johnson16-2-4	}&	10			&	10	&	0.00	&	&	10	&	0.00	\\
\tabincell{c}{	johnson32-2-4	}&	-			&	21	&	0.02	&	&	21	&	0.00	\\
\tabincell{c}{	johnson8-2-4	}&	5			&	5	&	0.00	&	&	5	&	0.00	\\
\tabincell{c}{	johnson8-4-4	}&	14			&	14	&	0.00	&	&	14	&	0.02	\\
\tabincell{c}{	keller4	}&	15			&	15	&	0.00	&	&	15	&	0.01	\\
\tabincell{c}{	keller5	}&	-			&	31	&	0.09	&	&	31	&	0.10	\\
\tabincell{c}{	keller6	}&	-			&	63	&	3.60	&	&	63	&	3.19	\\
\tabincell{c}{	MANN\_a27	}&	236			&	236 (235.9)	&	12.64	&	&	236	&	0.13	\\
\tabincell{c}{	MANN\_a45	}&	662			&	662 (661.4)	&	5.46	&	&	662	&	6.61	\\
\tabincell{c}{	MANN\_a81	}&	-			&	2162 (2113.9)	&	139.35	&	&2162		&12.33		\\
\tabincell{c}{	MANN\_a9	}&	26			&	26	&	0.00	&	&	26	&	0.01	\\
\tabincell{c}{	p\_hat1000-1	}&	-			&	13	&	0.28	&	&	13	&	0.08	\\
\tabincell{c}{	p\_hat1000-2	}&	-			&	56	&	0.43	&	&	56	&	0.32	\\
\tabincell{c}{	p\_hat1000-3	}&	-			&	82	&	0.33	&	&	82	&	0.17	\\
\tabincell{c}{	p\_hat1500-1	}&	-			&	14	&	1.46	&	&	14	&	0.33	\\
\tabincell{c}{	p\_hat1500-2	}&	-			&	80	&	1.75	&	&	80	&	0.95	\\
\tabincell{c}{	p\_hat1500-3	}&	-			&	114	&	0.61	&	&	114	&	0.16	\\
\tabincell{c}{	p\_hat300-1	}&	10			&	10	&	0.00	&	&	10	&	0.02	\\
\tabincell{c}{	p\_hat300-2	}&	30			&	30	&	0.01	&	&	30	&	0.03	\\
\tabincell{c}{	p\_hat300-3	}&	43			&	44	&	0.07	&	&	44	&	0.05	\\
\tabincell{c}{	p\_hat500-1	}&	12			&	12	&	0.06	&	&	12	&	0.03	\\
\tabincell{c}{	p\_hat500-2	}&	-			&	42	&	0.02	&	&	42	&	0.02	\\
\tabincell{c}{	p\_hat500-3	}&	-			&	62	&	0.18	&	&	62	&	0.06	\\
\tabincell{c}{	p\_hat700-1	}&	13			&	13	&	0.06	&	&	13	&	0.07	\\
\tabincell{c}{	p\_hat700-2	}&	50			&	52	&	0.06	&	&	52	&	0.03	\\
\tabincell{c}{	p\_hat700-3	}&	73			&	76	&	0.54	&	&	76	&	0.14	\\
\tabincell{c}{	san1000	}&	-			&	17	&	9.39	&	&	\textbf{18 (16.75)}	&	50.68	\\
\tabincell{c}{	san200\_0.7\_1	}&	-			&	31	&	0.59	&	&	31	&	0.05	\\
\tabincell{c}{	san200\_0.7\_2	}&	24			&	26 (25.4)	&	9.74	&	&	26	&	0.10	\\
\tabincell{c}{	san200\_0.9\_1	}&	90			&	90	&	0.01	&	&	90	&	0.01	\\
\tabincell{c}{	san200\_0.9\_2	}&	-			&	71	&	2.47	&	&	71 (69.35)	&	2.03	\\
\tabincell{c}{	san200\_0.9\_3	}&	-			&	54 (53.95)	&	72.31	&	&	54	&	7.16	\\
\tabincell{c}{	san400\_0.5\_1	}&	-			&	15	&	1.24	&	&	15 (14.9)	&	9.08	\\
\tabincell{c}{	san400\_0.7\_1	}&	-			&	41	&	0.20	&	&	\textbf{42 (41.55)}	&	10.27	\\
\tabincell{c}{	san400\_0.7\_2	}&	-			&	32	&	7.22	&	&	\textbf{33 (32.45)}	&	36.36	\\
\tabincell{c}{	san400\_0.7\_3	}&	-			&	27 (26.3)	&	12.27	&	&	\textbf{28 (27.75)}	&	66.10	\\
\tabincell{c}{	san400\_0.9\_1	}&	-			&	102 (101.3)	&	9.09	&	&	\textbf{103 (102.6)}	&	39.24	\\
\tabincell{c}{	sanr200\_0.7	}&	-			&	22	&	0.01	&	&	22	&	0.04	\\
\tabincell{c}{	sanr200\_0.9	}&	-			&	51	&	0.61	&	&	51	&	0.17	\\
\tabincell{c}{	sanr400\_0.5	}&	-			&	15	&	0.02	&	&	15	&	0.03	\\
\tabincell{c}{	sanr400\_0.7	}&	-			&	26	&	1.03	&	&	26	&	0.34	\\
\hline
\end{tabular}
\end{scriptsize}
\end{center}
\end{table}
\renewcommand{\baselinestretch}{1.0}\large\normalsize

\renewcommand{\baselinestretch}{0.8}\large\normalsize
\begin{table}[htp]
\begin{center}
\begin{scriptsize}
\caption{Comparisons of BLS-RLE for the $k$-plex when $k$=2}
\label{table2}
\begin{tabular}{lrrrrrrr}
\hline
\tabincell{c}{Benchmark set} &BEV &  &\multicolumn{2}{c}{FD-TS} &&\multicolumn{2}{c}{BLS-RLE} \\
\cline{2-2} \cline{4-5} \cline{7-8}
 &\#max   &  & \#max & \#avg.    & & \#max & \#avg.   \\
\hline
\tabincell{c}{	brock (12 instances)	}&	3	&	&	12	&	10	&	&	\emph{11}	&	10	\\
\tabincell{c}{	C (7 instances)	}&	0	&	&	3	&	3	&	&	\textbf{7}	&	3	\\
\tabincell{c}{	c-fat (7 instances)	}&	7	&	&	7	&	7	&	&	7	&	\emph{0}	\\
\tabincell{c}{	DSJC (2 instances)	}&	0	&	&	2	&	2	&	&	2	&	2	\\
\tabincell{c}{	gen (5 instances)	}&	0	&	&	3	&	2	&	&	\textbf{5}	&	\textbf{3}	\\
\tabincell{c}{	hamming (6 instances)	}&	5	&	&	6	&	6	&	&	6	&	6	\\
\tabincell{c}{	johnson (4 instances)	}&	3	&	&	4	&	4	&	&	4	&	4	\\
\tabincell{c}{	keller (3 instances)	}&	1	&	&	3	&	3	&	&	3	&	3	\\
\tabincell{c}{	MANN (4 instances)	}&	3	&	&	4	&	1	&	&	4	&	\textbf{4}	\\
\tabincell{c}{	phat (15 instances)	}&	4	&	&	15	&	15	&	&	15	&	15	\\
\tabincell{c}{	san (11 instances)	}&	1	&	&	6	&	4	&	&	\textbf{11}	&	4	\\
\tabincell{c}{	sanr (4 instances)	}&	0	&	&	4	&	4	&	&	4	&	4	\\
\hline
\end{tabular}
\end{scriptsize}
\end{center}
\end{table}
\renewcommand{\baselinestretch}{1.0}\large\normalsize

\renewcommand{\baselinestretch}{0.8}\large\normalsize
\begin{table}[!htp]
\begin{center}
\begin{scriptsize}
\caption{Comparisons of BLS-RLE for the $k$-plex when $k$=3}
\label{table3}
\begin{tabular}{lrrrrrrr}
\hline
\tabincell{c}{Benchmark set} &BEV &  &\multicolumn{2}{c}{FD-TS} &&\multicolumn{2}{c}{BLS-RLE} \\
\cline{2-2} \cline{4-5} \cline{7-8}
 &\#Max   &  & \#Max & \#Avg.    & & \#Max & \#Avg.   \\
\hline
\tabincell{c}{	brock (12 instances)	}&	1	&	&	11	&	9	&	&	\textbf{12}	&	9	\\
\tabincell{c}{	C (7 instances)	}&	0	&	&	3	&	2	&	&	\textbf{7}	&	\textbf{3}	\\
\tabincell{c}{	c-fat (7 instances)	}&	7	&	&	7	&	7	&	&	7	&	\emph{0}	\\
\tabincell{c}{	DSJC (2 instances)	}&	0	&	&	2	&	2	&	&	2	&	2	\\
\tabincell{c}{	gen (5 instances)	}&	0	&	&	5	&	4	&	&	5	&	\emph{2}	\\
\tabincell{c}{	hamming (6 instances)	}&	5	&	&	6	&	6	&	&	6	&	6	\\
\tabincell{c}{	johnson (4 instances)	}&	3	&	&	4	&	4	&	&	4	&	4	\\
\tabincell{c}{	keller (3 instances)	}&	1	&	&	2	&	2	&	&	\textbf{3}	&	\textbf{3}	\\
\tabincell{c}{	MANN (4 instances)	}&	3	&	&	4	&	3	&	&	4	&	\textbf{4}	\\
\tabincell{c}{	phat (15 instances)	}&	2	&	&	15	&	15	&	&	15	&	15	\\
\tabincell{c}{	san (11 instances)	}&	1	&	&	10	&	8	&	&	\textbf{11}	&	\emph{2}	\\
\tabincell{c}{	sanr (4 instances)	}&	0	&	&	4	&	4	&	&	4	&	4	\\
\hline
\end{tabular}
\end{scriptsize}
\end{center}
\end{table}
\renewcommand{\baselinestretch}{1.0}\large\normalsize

\renewcommand{\baselinestretch}{0.8}\large\normalsize
\begin{table}[!htp]
\begin{center}
\begin{scriptsize}
\caption{Comparisons of BLS-RLE for the $k$-plex when $k$=4}
\label{table4}
\begin{tabular}{lrrrrrrr}
\hline
\tabincell{c}{Benchmark set} &BEV &  &\multicolumn{2}{c}{FD-TS} &&\multicolumn{2}{c}{BLS-RLE} \\
\cline{2-2} \cline{4-5} \cline{7-8}
 &\#Max   &  & \#Max & \#Avg.    & & \#Max & \#Avg.   \\
\hline
\tabincell{c}{	brock (12 instances)	}&	0	&	&	11	&	8	&	&	\textbf{12}	&	\textbf{9}	\\
\tabincell{c}{	C (7 instances)	}&	0	&	&	2	&	2	&	&	\textbf{7}	&	2	\\
\tabincell{c}{	c-fat (7 instances)	}&	7	&	&	7	&	7	&	&	7	&	\emph{0}	\\
\tabincell{c}{	DSJC (2 instances)	}&	0	&	&	2	&	1	&	&	2	&	1	\\
\tabincell{c}{	gen (5 instances)	}&	0	&	&	5	&	5	&	&	5	&	\emph{2}	\\
\tabincell{c}{	hamming (6 instances)	}&	3	&	&	6	&	5	&	&	\emph{5}	&	\emph{4}	\\
\tabincell{c}{	johnson (4 instances)	}&	3	&	&	4	&	4	&	&	4	&	4	\\
\tabincell{c}{	keller (3 instances)	}&	0	&	&	2	&	1	&	&	\textbf{3}	&	1	\\
\tabincell{c}{	MANN (4 instances)	}&	3	&	&	4	&	3	&	&	4	&	\textbf{4}	\\
\tabincell{c}{	phat (15 instances)	}&	1	&	&	15	&	14	&	&	15	&	\textbf{15}	\\
\tabincell{c}{	san (11 instances)	}&	1	&	&	11	&	9	&	&	11	&	\emph{2}	\\
\tabincell{c}{	sanr (4 instances)	}&	0	&	&	4	&	4	&	&	4	&	4	\\
\hline
\end{tabular}
\end{scriptsize}
\end{center}
\end{table}
\renewcommand{\baselinestretch}{1.0}\large\normalsize

\renewcommand{\baselinestretch}{0.8}\large\normalsize
\begin{table}[!htb]
\begin{center}
\begin{scriptsize}
\caption{Comparisons of BLS-RLE for the $k$-plex when $k$=5}
\label{table5}
\begin{tabular}{lrrrrrrr}
\hline
\tabincell{c}{Benchmark set} &BEV &  &\multicolumn{2}{c}{FD-TS} &&\multicolumn{2}{c}{BLS-RLE} \\
\cline{2-2} \cline{4-5} \cline{7-8}
 &\#Max   &  & \#Max & \#Avg.    & & \#Max & \#Avg.   \\
\hline
\tabincell{c}{	brock (12 instances)	}&	0	&	&	12	&	8	&	&	12	&	\textbf{11}	\\
\tabincell{c}{	C (7 instances)	}&	0	&	&	3	&	2	&	&	\textbf{7}	&	\textbf{3}	\\
\tabincell{c}{	c-fat (7 instances)	}&	7	&	&	7	&	7	&	&	7	&	\emph{1}	\\
\tabincell{c}{	DSJC (2 instances)	}&	0	&	&	2	&	1	&	&	2	&	1	\\
\tabincell{c}{	gen (5 instances)	}&	0	&	&	5	&	5	&	&	5	&	\emph{2}	\\
\tabincell{c}{	hamming (6 instances)	}&	2	&	&	6	&	4	&	&	6	&	4	\\
\tabincell{c}{	johnson (4 instances)	}&	1	&	&	4	&	4	&	&	4	&	4	\\
\tabincell{c}{	keller (3 instances)	}&	0	&	&	3	&	2	&	&	3	&	\emph{1}	\\
\tabincell{c}{	MANN (4 instances)	}&	2	&	&	3	&	3	&	&	\textbf{4}	&	\textbf{4}	\\
\tabincell{c}{	phat (15 instances)	}&	0	&	&	13	&	12	&	&	\textbf{15}	&	12	\\
\tabincell{c}{	san (11 instances)	}&	1	&	&	11	&	10	&	&	\emph{10}	&	\emph{6}	\\
\tabincell{c}{	sanr (4 instances)	}&	0	&	&	4	&	4	&	&	4	&	4	\\
\hline
\end{tabular}
\end{scriptsize}
\end{center}
\end{table}
\renewcommand{\baselinestretch}{1.0}\large\normalsize

\renewcommand{\baselinestretch}{0.9}\large\normalsize
\begin{table}[!htb]
\begin{scriptsize}
\begin{center}
\centering
\caption{New best results for DIMACS instances by BLS-RLE}
\label{tableNewBest}
\begin{tabular}{l|l|rrcl|l|rr}
\cline{1-4}	\cline{6-9}	
 & Instance &Max$_{pre}$& Max  & & & Instance &Max$_{pre}$ & Max \\
\cline{1-4}	\cline{6-9}	
$k$ = 2 &	C1000.9	& 81  &	82	&	&	$k$ = 4 & brock800\_4	& 33      &	34 \\		
&	C2000.5	& 19    &	20	&	&	& C1000.9	& 107       &	109 \\		
&	C2000.9	& 90    &	93	&	&	& C2000.5	& 25       &	26	\\	
&	C4000.5	& 20    &	21	&	&	& C2000.9	& 118       &	123	\\	
&	gen400\_p0.9\_65	& 73       &	74	&	&	&	C4000.5	& 26       &	27	\\
&	gen400\_p0.9\_75	& 79       &	80	&	&	&	C500.9	& 92        &	93	\\
&	san1000	&  17       &	18	&	&	&	keller6	&  107      &	112	\\
\cline{6-9}										
&	san400\_0.7\_1	&  41    &	42	&	&	$k$ = 5 &	C1000.9	&  119      &	122	\\
&	san400\_0.7\_2	&  32    &	33	&	&	&	C2000.9	& 132          &	137	\\
&	san400\_0.7\_3	&  27    &	28	&	&	&	C4000.5	& 29       &	30	\\
&	san400\_0.9\_1	&  102    &	103	&	&	&	C500.9	& 103       &	104	\\
\cline{1-4}										
$k$ = 3 & brock800\_1	&     &	30		&	&	&	MANN\_a81	& 3135        &	3240	\\
& C1000.9	&  95       &	96		&	&	& p\_hat1500-3	&   164       &	165	\\	
& C2000.5	&  22       &	23		&	&	& p\_hat500-3	&   89        &	90	\\	
\cline{6-9}	
& C2000.9	&  105       &	109		&	&	\multicolumn{1}{c}{} & \multicolumn{1}{c}{}	&	&	\\	
& C4000.5	&  23       &	24		&	&\multicolumn{1}{c}{}	& \multicolumn{1}{c}{}	&	&	\\	
& keller6	&  90       &	93		&	&\multicolumn{1}{c}{}	& \multicolumn{1}{c}{}	&	&	\\	
& san400\_0.7\_3	& 38     & 39			&	&\multicolumn{1}{c}{} &	\multicolumn{1}{c}{}	&	&	\\	
\cline{1-4}		
\end{tabular}
\end{center}
\end{scriptsize}
\end{table}
\renewcommand{\baselinestretch}{1.0}\large\normalsize

\emph{BEV} lists the best known results obtained by the four reference exact algorithms of \cite{balasundaram2011clique,mcclosky2012combinatorial,moser2012exact,trukhanov2013algorithms}. Each of these exact algorithms terminates after a maximum of 3 hours run time (except the algorithm in \cite{mcclosky2012combinatorial} which terminates after 1 hour). We acknowledge that the experimental platforms used by these methods vary, however the large difference in running times means that the impact of differing CPU speeds on results is nominal. As such we have chosen to extract the computational results of the exact algorithms from the corresponding papers directly. FD-TS gives the results of the heuristic method presented in \cite{zhou2017frequency}, which uses the same termination criteria as our method, stopping after 180 seconds. %``BLS-RLE" presents the computational results of our proposed BLS-RLE, and the stopping condition is the same as the stopping condition used in FD-TS.
For FD-TS and BLS-RLE, the best result and average results that are achieved over 20 runs are given. The \emph{time} column provides the average time in seconds required to find the best solution by those runs that obtained the best result.

From this table, we observe that the reference exact algorithms are able to solve a subset of the 80 instances to optimality, however these consist mainly of small instances. In general, both the reference heuristic FD-TS and BLS-RLE can obtain good results for these instances in a short period of time. In most cases, the average performance matches the best performance, achieving the same results over all 20 runs. New best known solutions are found for eleven of the 80 instances: C1000.9, C2000.5, C2000.9, C4000.5, gen400\_p0.9\_65, gen400\_p0.9\_75, san1000, san400\_0.7\_1, san400\_0.7\_2, san400\_0.7\_3 and san400\_0.9\_1. Interestingly all of these instances are in the minority, where the average performance and best performance differ. It may be the case that these instances are intrinsically more difficult than some of the others in the set, indeed the average time to find a best solution is also generally longer for these cases.

Table \ref{table3}, \ref{table4} and \ref{table5} summarize the computational results of BLS-RLE for the $k$-plex when $k$ = 3, 4, and 5 respectively. Each cell in this column represents the number of instances in that subset where the best known value was found. Here, the $Avg.$ represents the average result over the number of runs by the reference algorithm. For BLS-RLE, results are highlighted in bold in the case that BLS-RLE outperforms FD-TS, no highlighting is applied when the results are the same, and results are italic when FD-TS outperforms BLS-RLE.

Over all 80 DIMACS instances, the BLS-RLE algorithm can match the best-known results for the $k$-plex when $k$ = 2, 3, 4, 5 in all but four cases. In particular, BLS-RLE improves the best known results for 32 instances, finding eleven new best solutions for $k$ = 2 (in bold in Table \ref{tabledet2}) and seven new best solutions for $k$ = 3, 4, and 5. This comparison shows that BLS-RLE offers highly competitive performance for these problem instances. Table \ref{tableNewBest} provides the specific results for the 32 instances where a new best solution has been found.

\section{Analysis on the reinforcement learning}
\label{SecAnalysis}

Table \ref{tableRND} provides a direct comparison between BLS-RLE and BLS-RND (where $l$, $e$ and $b$ are selected randomly at each step, but share the same range with BLS-RLE), over all 80 benchmark instances. The number of instances in each benchmark set is provided in parenthesis. Despite its simplicity, determining the perturbation operator and perturbation magnitude is clearly an effective strategy in this problem domain, often matching or beating BLS-RLE in terms of both maximum and average results obtained. This performance is not necessarily linked to $k$, with both methods seemingly scaling in a similar manner as the value of $k$ increases. Nevertheless, BLS-RLE performs better than BLS-RND on the hard instances, i.e, BLS-RLE can find the new results 34, 26 for brock800-4 and C2000.5 while BLS-RND can only achieve the best-known results 33 and 25 when $k$ = 4.

\renewcommand{\baselinestretch}{0.8}\large\normalsize
\begin{table}[htp]
\begin{center}
\begin{scriptsize}
\caption{Comparison between BLS-RLE and BLS-RND over 80 DIMACS $k$-plex instances}
\label{tableRND}
%\resizebox{\textwidth}{15mm}{
\begin{tabular}{lrrrrrrrrrrr}
\hline
 &\multicolumn{5}{c}{k=2} &&\multicolumn{5}{c}{k=3} \\
\cline{2-6} \cline{8-12}
\tabincell{c}{Benchmark set}&\multicolumn{2}{c}{BLS-RLE} &&\multicolumn{2}{c}{BLS-RND} &&\multicolumn{2}{c}{BLS-RLE} &&\multicolumn{2}{c}{BLS-RND} \\
\cline{2-3} \cline{5-6} \cline{8-9} \cline{11-12}
  &  \#max & \#avg.     && \#max & \#avg.  & & \#max & \#avg.  && \#max & \#avg. \\
\hline
brock (12)	&	11	&	10	&&	11	&	10	&&	12	&	9	&&	12	&	9	\\
C (7)	&	7	&	3	&&	7	&	4	&&	\textbf{7}	&	3	&&	6	&	3	\\
\tabincell{c}{	c-fat (7)	}&	7	&	0	&&	7	&	0	&&	7	&	0	&&	7	&	0 \\
\tabincell{c}{	DSJC (2)	}&	2	&	2	&&	2	&	2	&&	2	&	2	&&	2	&	2 \\
\tabincell{c}{	gen (5)	}&	5	&	3	&&	5	&	3	&&	5	&	2	&&	5	&	2	\\
\tabincell{c}{	hamming (7)	}&	6	&	6	&&	6	&	6	&&	6	&	6	&&	6	&	6	\\
\tabincell{c}{	johnson (4)	}&	4	&	4	&&	4	&	4	&&	4	&	4	&&	4	&	4	\\
\tabincell{c}{	keller (3)	}&	3	&	3	&&	3	&	3	&&	3	&	\textbf{3}	&&	3	&	2	\\
\tabincell{c}{	MANN (4)	}&	4	&	4	&&	4	&	4	&&	4	&	4	&&	4	&	4	\\
\tabincell{c}{	phat (15)	}&	15	&	15	&&	15	&	15	&&	15	&	15	&&	15	&	15	\\
\tabincell{c}{	san (11)	}&	10	&	4	&&	10	&	4	&&	11	&	2	&&	11	&	\textbf{3} \\
\tabincell{c}{	sanr (4)	}&	4	&	4	&&	4	&	4	&&	4	&	4	&&	4	&	4	\\
\hline
\end{tabular}
%}
%\resizebox{\textwidth}{15mm}{
\begin{tabular}{lrrrrrrrrrrrr}
\hline
 &\multicolumn{5}{c}{k=4} &&\multicolumn{5}{c}{k=5} \\
\cline{2-6} \cline{8-12}
\tabincell{c}{Benchmark set}&\multicolumn{2}{c}{BLS-RLE} &&\multicolumn{2}{c}{BLS-RND} &&\multicolumn{2}{c}{BLS-RLE} &&\multicolumn{2}{c}{BLS-RND} \\
\cline{2-3} \cline{5-6} \cline{8-9} \cline{11-12}
  &  \#max & \#avg.     && \#max & \#avg.  & & \#max & \#avg.  && \#max & \#avg. \\
\hline
brock (12)	&	\textbf{12}	&	9	&&	11	&	\textbf{10}	&&	12	&	\textbf{11}	&&	12	&	10	\\
C (7)	&	\textbf{7}	&	2	&&	6	&	2	&&	7	&	3	&&	7	&	3	\\
\tabincell{c}{	c-fat (7)	} &	7	&	0	&&	7	&	0	&&	7	&	1	&&	7	&	1	\\
\tabincell{c}{	DSJC (2)	} &	2	&	1	&&	2	&	1	&&	2	&	1	&&	2	&	1	\\
\tabincell{c}{	gen (5)	}     &	5	&	2	&&	5	&	2	&&	5	&	2	&&	5	&	2	\\
\tabincell{c}{	hamming (7)	} &	5	&	4	&&	5	&	\textbf{5}	&&	6	&	\textbf{4}	&&	6	&	3	\\
\tabincell{c}{	johnson (4)	} &	4	&	4	&&	4	&	4	&&	4	&	4	&&	4	&	4	\\
\tabincell{c}{	keller (3)	} &	3	&	1	&&	3	&	1	&&	3	&	1	&&	3	&	1	\\
\tabincell{c}{	MANN (4)	} &	3	&	3	&&	4	&	4	&&	4	&	4	&&	4	&	4	\\
\tabincell{c}{	phat (15)	} &	15	&	15	&&	15	&	15	&&	15	&	12	&&	15	&	\textbf{14}	\\
\tabincell{c}{	san (11)	} &	\textbf{11}	&	2	&&	10	&	\textbf{3}	&&	10	&	6	&&	10	&	6	\\
\tabincell{c}{	sanr (4)	} &	4	&	4	&&	4	&	4	&&	4	&	4	&&	4	&	4	\\
\hline
\end{tabular}
%}
\end{scriptsize}
\end{center}
\end{table}
\renewcommand{\baselinestretch}{1.0}\large\normalsize

\section{Conclusion}
\label{Sec_Conclusion}
The BLS-RLE method presented in this paper is the first study for the maximum $k$-plex problem focusing on a cooperative approach between a local search procedure and a reinforcement learning strategy. BLS-RLE alternates between an intensification stage with descent local search and a diversification stage with directed or random perturbations. A reinforcement learning mechanism is employed to interdependently control three parameters, the probability $e$ of using a particular type of perturbation operator, the degree of random perturbation $b$ and the number of perturbation moves $l$ used in order to escape local optima traps. A novel strategy for enumerating possible values for $l$ and a new parameter control for generating the triples $(l,e,b)$ have been proposed, differing from the existing approaches in the literature.

Experimental evaluations over 80 benchmark instances for $k$ = 2, 3, 4, 5 showed that the proposed BLS-RLE algorithm is highly competitive in comparison with state-of-the-art exact and heuristic algorithms for the maximum $k$-plex problem. In particular, BLS-RLE can improve the best known results for 32 instances. Although BLS-RLE has shown to be competitive when compared to existing approaches, random sampling of parameters has also shown promising performance (BLS-RND), with the added benefit of reducing the number of design choices that are required and reducing the burden of parameter tuning. Our current work in this area is focused on using different action selection models, comparing the ability of different approaches to learn good parameter triples. In future we will examine other automated parameter tuning methods, such as \emph{irace} \cite{lopez2016irace}, to manage the parameter combinations that are considered during the search process.

\section{Appendix}
\renewcommand{\baselinestretch}{0.5}\large\normalsize
\begin{table}
\begin{center}
\begin{scriptsize}
\caption{Comparisons of BLS-RLE for the $k$-plex when $k$=3}
\label{tabledet3}
%\begin{tabular}{p{1.1cm}p{0.4cm}p{0.4cm}p{1.1cm}p{0.4cm}p{1.1cm}p{0.3cm}p{0.0cm}}
\begin{tabular}{lrrrrrr}
\hline
\tabincell{c}{Instance} &BKV  &\multicolumn{2}{c}{FD-TS} &&\multicolumn{2}{c}{BLS-RLE} \\
\cline{3-4} \cline{6-7}
 &    & Max(Avg.) & time(s)    & & Max(Avg.) & time(s)   \\
\hline
\tabincell{c}{	brock200\_1	}&	24	&	30	&	0.05 	&	&	30	&	0.028	\\
\tabincell{c}{	brock200\_2	}&	16	&	16	&	0.27 	&	&	16	&	0.112	\\
\tabincell{c}{	brock200\_3	}&	19	&	20	&	0.02 	&	&	20	&	0.0235	\\
\tabincell{c}{	brock200\_4	}&	20	&	23	&	0.07 	&	&	23	&	0.05	\\
\tabincell{c}{	brock400\_1	}&	23	&	36	&	7.24 	&	&	36	&	2.0655	\\
\tabincell{c}{	brock400\_2	}&	27	&	36	&	15.37 	&	&	36	&	2.635	\\
\tabincell{c}{	brock400\_3	}&	-	&	36	&	6.59 	&	&	36	&	0.957	\\
\tabincell{c}{	brock400\_4	}&	27	&	36	&	4.60 	&	&	36	&	0.752	\\
\tabincell{c}{	brock800\_1	}&	-	&	29	&	12.35 	&	&	\textbf{30(29.05)}	&	9.6445	\\
\tabincell{c}{	brock800\_2	}&	-	&	30(29.30)	&	31.20 	&	&	30(29.95)	&	50.016	\\
\tabincell{c}{	brock800\_3	}&	-	&	30(29.20)	&	11.71 	&	&	30(29.95)	&	55.7225	\\
\tabincell{c}{	brock800\_4	}&	-	&	29	&	14.35 	&	&	29	&	1.0425	\\
\tabincell{c}{	C1000.9	}&	-	&	95(93.75)	&	58.59 	&	&	\textbf{96(95.55)}	&	54.1065	\\
\tabincell{c}{	C125.9	}&	-	&	51	&	1.89 	&	&	51	&	0.4015	\\
\tabincell{c}{	C2000.5	}&	-	&	22(21.90)	&	62.35 	&	&	\textbf{23(22.05)}	&	10.8295	\\
\tabincell{c}{	C2000.9	}&	-	&	105(103.40)	&	69.41 	&	&	\textbf{109(107.3)}	&	69.554	\\
\tabincell{c}{	C250.9	}&	-	&	65	&	22.29 	&	&	65	&	9.86	\\
\tabincell{c}{	C4000.5	}&	-	&	23	&	69.37 	&	&	\textbf{24(23.3)}	&	21.172	\\
\tabincell{c}{	C500.9	}&	-	&	81(80.95)	&	57.72 	&	&	81	&	6.72	\\
\tabincell{c}{	c-fat200-1	}&	12	&	12	&	0.00 	&	&	12(10.95)	&	0.0095	\\
\tabincell{c}{	c-fat200-2	}&	24	&	24	&	0.01 	&	&	24(22.3)	&	0.0125	\\
\tabincell{c}{	c-fat200-5	}&	58	&	58	&	0.01 	&	&	58(56.85)	&	0.0135	\\
\tabincell{c}{	c-fat500-10	}&	126	&	126	&	0.22 	&	&	126(124.9)	&	0.018	\\
\tabincell{c}{	c-fat500-1	}&	14	&	14	&	0.00 	&	&	14(12.4)	&	0.021	\\
\tabincell{c}{	c-fat500-2	}&	26	&	26	&	0.00 	&	&	26(25.25)	&	0.0215	\\
\tabincell{c}{	c-fat500-5	}&	64	&	64	&	0.02 	&	&	64(62.45)	&	0.022	\\
\tabincell{c}{	DSJC1000\_5	}&	-	&	21	&	25.35 	&	&	21	&	8.4675	\\
\tabincell{c}{	DSJC500\_5	}&	-	&	19	&	2.81 	&	&	19	&	0.6185	\\
\tabincell{c}{	gen200\_p0.9\_44	}&	-	&	66	&	0.09 	&	&	66	&	0.0365	\\
\tabincell{c}{	gen200\_p0.9\_55	}&	-	&	64	&	0.14 	&	&	64	&	0.0815	\\
\tabincell{c}{	gen400\_p0.9\_55	}&	-	&	87	&	26.62 	&	&	87(86.2)	&	8.13	\\
\tabincell{c}{	gen400\_p0.9\_65	}&	-	&	101(100.45)	&	10.68 	&	&	101(99.35)	&	0.457	\\
\tabincell{c}{	gen400\_p0.9\_75	}&	-	&	114	&	0.35 	&	&	114(112.5)	&	0.185	\\
\tabincell{c}{	hamming10-2	}&	512	&	512	&	4.66 	&	&	512	&	2.2325	\\
\tabincell{c}{	hamming10-4	}&	46	&	64	&	1.13 	&	&	64	&	0.4625	\\
\tabincell{c}{	hamming6-2	}&	32	&	32	&	0.00 	&	&	32	&	0.0215	\\
\tabincell{c}{	hamming6-4	}&	8	&	8	&	0.00 	&	&	8	&	0.0185	\\
\tabincell{c}{	hamming8-2	}&	128	&	128	&	0.09 	&	&	128	&	0.0185	\\
\tabincell{c}{	hamming8-4	}&	20	&	20	&	0.01 	&	&	20	&	0.015	\\
\tabincell{c}{	johnson16-2-4	}&	16	&	16	&	0.00 	&	&	16	&	0.0045	\\
\tabincell{c}{	johnson32-2-4	}&	-	&	32	&	0.05 	&	&	32	&	0.016	\\
\tabincell{c}{	johnson8-2-4	}&	8	&	8	&	0.00 	&	&	8	&	0.007	\\
\tabincell{c}{	johnson8-4-4	}&	18	&	18	&	0.00 	&	&	18	&	0.02	\\
\tabincell{c}{	keller4	}&	21	&	21	&	0.09 	&	&	21	&	0.015	\\
\tabincell{c}{	keller5	}&	-	&	45	&	8.19 	&	&	45	&	1.361	\\
\tabincell{c}{	keller6	}&	-	&	90(87.80)	&	66.21 	&	&	\textbf{93}	&	47.53	\\
\tabincell{c}{	MANN\_a27	}&	351	&	351	&	0.41 	&	&	351	&	0.049	\\
\tabincell{c}{	MANN\_a45	}&	990	&	990	&	7.40 	&	&	990	&	0.84	\\
\tabincell{c}{	MANN\_a81	}&	-	&	3240(3125.35)	&	138.36 	&	&3240		&31.3		\\
\tabincell{c}{	MANN\_a9	}&	36	&	36	&	0.00 	&	&	36	&	0.007	\\
\tabincell{c}{	p\_hat1000-1	}&	-	&	15	&	0.15 	&	&	15	&	0.038	\\
\tabincell{c}{	p\_hat1000-2	}&	-	&	67	&	0.81 	&	&	67	&	0.118	\\
\tabincell{c}{	p\_hat1000-3	}&	-	&	98	&	2.19 	&	&	98	&	0.9785	\\
\tabincell{c}{	p\_hat1500-1	}&	-	&	17	&	22.47 	&	&	17	&	6.555	\\
\tabincell{c}{	p\_hat1500-2	}&	-	&	93	&	0.41 	&	&	93	&	0.115	\\
\tabincell{c}{	p\_hat1500-3	}&	-	&	133	&	17.85 	&	&	133	&	0.7665	\\
\tabincell{c}{	p\_hat300-1	}&	12	&	12	&	0.00 	&	&	12	&	0.0095	\\
\tabincell{c}{	p\_hat300-2	}&	30	&	36	&	0.02 	&	&	36	&	0.0315	\\
\tabincell{c}{	p\_hat300-3	}&	43	&	52	&	0.06 	&	&	52	&	0.0805	\\
\tabincell{c}{	p\_hat500-1	}&	14	&	14	&	0.20 	&	&	14	&	0.142	\\
\tabincell{c}{	p\_hat500-2	}&	-	&	50	&	0.12 	&	&	50	&	0.062	\\
\tabincell{c}{	p\_hat500-3	}&	-	&	72	&	1.24 	&	&	72	&	0.2565	\\
\tabincell{c}{	p\_hat700-1	}&	13	&	15	&	0.13 	&	&	15	&	0.0625	\\
\tabincell{c}{	p\_hat700-2	}&	50	&	62	&	1.35 	&	&	62	&	0.327	\\
\tabincell{c}{	p\_hat700-3	}&	73	&	89	&	2.13 	&	&	89	&	0.163	\\
\tabincell{c}{	san1000	}&	-	&	25	&	6.73 	&	&	25(23.4)	&	3.7185	\\
\tabincell{c}{	san200\_0.7\_1	}&	-	&	46(45.70)	&	1.51 	&	&	46(45.7)	&	0.7615	\\
\tabincell{c}{	san200\_0.7\_2	}&	36	&	37	&	0.21 	&	&	37	&	0.107	\\
\tabincell{c}{	san200\_0.9\_1	}&	125	&	125	&	0.02 	&	&	125(122.5)	&	0.02	\\
\tabincell{c}{	san200\_0.9\_2	}&	-	&	105	&	0.02 	&	&	105(102)	&	0.011	\\
\tabincell{c}{	san200\_0.9\_3	}&	-	&	73	&	7.48 	&	&	73(69.5)	&	11.0195	\\
\tabincell{c}{	san400\_0.5\_1	}&	-	&	22	&	3.61 	&	&	22(20.75)	&	24.101	\\
\tabincell{c}{	san400\_0.7\_1	}&	-	&	61	&	2.49 	&	&	61(60.9)	&	60.1145	\\
\tabincell{c}{	san400\_0.7\_2	}&	-	&	47(46.10)	&	0.46 	&	&	47(46.8)	&	38.1925	\\
\tabincell{c}{	san400\_0.7\_3	}&	-	&	38	&	11.17 	&	&	\textbf{39(37.35)}	&	21.7735	\\
\tabincell{c}{	san400\_0.9\_1	}&	-	&	150	&	0.09 	&	&	150	&	0.0345	\\
\tabincell{c}{	sanr200\_0.7	}&	-	&	26	&	0.03 	&	&	26	&	0.039	\\
\tabincell{c}{	sanr200\_0.9	}&	-	&	61	&	2.25 	&	&	61	&	0.448	\\
\tabincell{c}{	sanr400\_0.5	}&	-	&	18	&	0.09 	&	&	18	&	0.021	\\
\tabincell{c}{	sanr400\_0.7	}&	-	&	30	&	0.45 	&	&	30	&	0.104	\\
\hline
\end{tabular}
\end{scriptsize}
\end{center}
\end{table}
\renewcommand{\baselinestretch}{1.0}\large\normalsize

\renewcommand{\baselinestretch}{0.5}\large\normalsize
\begin{table}
\begin{center}
\begin{scriptsize}
\caption{Comparisons of BLS-RLE for the $k$-plex when $k$=4}
\label{tabledet4}
\begin{tabular}{lrrrrrr}
\hline
\tabincell{c}{Instance} & \multicolumn{1}{c}{BKV}   &\multicolumn{2}{c}{FD-TS} &&\multicolumn{2}{c}{BLS-RLE} \\
\cline{3-4} \cline{6-7}
 &  &Max(Avg.) & time(s)    & & Max(Avg.) & time(s)   \\
\hline
\tabincell{c}{	brock200\_1	}&	27	&	35	&	3.59 	&	&	35	&	0.3945	\\
\tabincell{c}{	brock200\_2	}&	17	&	18	&	0.06 	&	&	18	&	0.061	\\
\tabincell{c}{	brock200\_3	}&	19	&	23	&	0.03 	&	&	23	&	0.027	\\
\tabincell{c}{	brock200\_4	}&	21	&	26	&	0.03 	&	&	26	&	0.0215	\\
\tabincell{c}{	brock400\_1	}&	23	&	41	&	41.37 	&	&	41	&	5.678	\\
\tabincell{c}{	brock400\_2	}&	29	&	41	&	33.51 	&	&	41	&	1.2835	\\
\tabincell{c}{	brock400\_3	}&		&	41	&	5.90 	&	&	41	&	0.376	\\
\tabincell{c}{	brock400\_4	}&	30	&	41	&	1.76 	&	&	41	&	0.4995	\\
\tabincell{c}{	brock800\_1	}&	-	&	34(33.20)	&	24.43 	&	&	34	&	30.1935	\\
\tabincell{c}{	brock800\_2	}&	-	&	34(33.15)	&	26.40 	&	&	34(33.75)	&	47.5185	\\
\tabincell{c}{	brock800\_3	}&	-	&	34(33.15)	&	28.46 	&	&	34(33.9)	&	47.9195	\\
\tabincell{c}{	brock800\_4	}&	-	&	33	&	32.04 	&	&	\textbf{34(33.05)}	&	5.788	\\
\tabincell{c}{	C1000.9	}&	-	&	107(106.00)	&	48.91 	&	&	\textbf{109(108.75)}	&	41.6625	\\
\tabincell{c}{	C125.9	}&	-	&	58	&	0.07 	&	&	58	&	0.022	\\
\tabincell{c}{	C2000.5	}&	-	&	25(24.50)	&	22.79 	&	&	\textbf{26(25.05)}	&	14.574	\\
\tabincell{c}{	C2000.9	}&	-	&	118(116.80)	&	63.85 	&	&	\textbf{123(121.6)}	&	63.8995	\\
\tabincell{c}{	C250.9	}&	-	&	75	&	4.81 	&	&	75	&	0.4745	\\
\tabincell{c}{	C4000.5	}&	-	&	26(25.55)	&	53.64 	&	&	\textbf{27(26.1)}	&	23.646	\\
\tabincell{c}{	C500.9	}&	-	&	92(91.75)	&	55.11 	&	&	\textbf{93(92.45)}	&	36.314	\\
\tabincell{c}{	c-fat200-1	}&	12	&	12	&	0.00 	&	&	12(11.5)	&	0.005	\\
\tabincell{c}{	c-fat200-2	}&	24	&	24	&	0.02 	&	&	24(22.3)	&	0.0165	\\
\tabincell{c}{	c-fat200-5	}&	58	&	58	&	0.00 	&	&	58(57.45)	&	0.0115	\\
\tabincell{c}{	c-fat500-10	}&	126	&	126	&	0.23 	&	&	126(125.45)	&	0.0465	\\
\tabincell{c}{	c-fat500-1	}&	14	&	14	&	0.00 	&	&	14(12.9)	&	0.0165	\\
\tabincell{c}{	c-fat500-2	}&	26	&	26	&	0.00 	&	&	26(25.2)	&	0.011	\\
\tabincell{c}{	c-fat500-5	}&	64	&	64	&	0.03 	&	&	64(62.5)	&	0.0305	\\
\tabincell{c}{	DSJC1000\_5	}&	-	&	24(23.05)	&	5.63 	&	&	24(23.4)	&	25.494	\\
\tabincell{c}{	DSJC500\_5	}&	-	&	21	&	0.12 	&	&	21	&	0.2195	\\
\tabincell{c}{	gen200\_p0.9\_44	}&	-	&	76	&	0.03 	&	&	76	&	0.0245	\\
\tabincell{c}{	gen200\_p0.9\_55	}&	-	&	73	&	0.12 	&	&	73	&	0.035	\\
\tabincell{c}{	gen400\_p0.9\_55	}&	-	&	112	&	0.19 	&	&	112(110.2)	&	0.1505	\\
\tabincell{c}{	gen400\_p0.9\_65	}&	-	&	132	&	0.25 	&	&	132(129.8)	&	0.0325	\\
\tabincell{c}{	gen400\_p0.9\_75	}&	-	&	136	&	0.10 	&	&	136(133.9)	&	0.039	\\
\tabincell{c}{	hamming10-2	}&	512	&	512	&	8.62 	&	&	512	&	1.499	\\
\tabincell{c}{	hamming10-4	}&	51	&	68(67.20)	&	20.64 	&	&	68(67.95)	&	53.0775	\\
\tabincell{c}{	hamming6-2	}&	40	&	40	&	0.00 	&	&	40	&	0.011	\\
\tabincell{c}{	hamming6-4	}&	10	&	10	&	0.00 	&	&	10	&	0.011	\\
\tabincell{c}{	hamming8-2	}&	128	&	129	&	22.41 	&	&	\emph{128}	&	0.0655	\\
\tabincell{c}{	hamming8-4	}&	20	&	25	&	0.28 	&	&	25	&	0.2625	\\
\tabincell{c}{	johnson16-2-4	}&	19	&	19	&	0.00 	&	&	19	&	0.0145	\\
\tabincell{c}{	johnson32-2-4	}&	-	&	38	&	0.38 	&	&	38	&	0.118	\\
\tabincell{c}{	johnson8-2-4	}&	9	&	9	&	0.00 	&	&	9	&	0.001	\\
\tabincell{c}{	johnson8-4-4	}&	22	&	22	&	0.00 	&	&	22	&	0.0115	\\
\tabincell{c}{	keller4	}&	22	&	23	&	0.06 	&	&	23	&	0.213	\\
\tabincell{c}{	keller5	}&	-	&	53(52.75)	&	53.67 	&	&	53(52.75)	&	49.822	\\
\tabincell{c}{	keller6	}&	-	&	107(103.45)	&	67.88 	&	&	\textbf{112(106.8)}	&	70.3805	\\
\tabincell{c}{	MANN\_a27	}&	351	&	351	&	0.45 	&	&	351	&	0.1055	\\
\tabincell{c}{	MANN\_a45	}&	990	&	990	&	7.48 	&	&	990	&	1.954	\\
\tabincell{c}{	MANN\_a81	}&	-	&	3240(2788.70)	&	147.51 	&	&3240		&29.2		\\
\tabincell{c}{	MANN\_a9	}&	36	&	36	&	0.00 	&	&	36	&	0	\\
\tabincell{c}{	p\_hat1000-1	}&	-	&	18	&	4.09 	&	&	18	&	1.9085	\\
\tabincell{c}{	p\_hat1000-2	}&	-	&	76	&	28.89 	&	&	76	&	0.5585	\\
\tabincell{c}{	p\_hat1000-3	}&	-	&	111	&	3.50 	&	&	111	&	0.311	\\
\tabincell{c}{	p\_hat1500-1	}&	-	&	19	&	3.73 	&	&	19	&	0.83	\\
\tabincell{c}{	p\_hat1500-2	}&	-	&	107(106.50)	&	29.46 	&	&	107	&	9.111	\\
\tabincell{c}{	p\_hat1500-3	}&	-	&	150	&	3.90 	&	&	150	&	0.542	\\
\tabincell{c}{	p\_hat300-1	}&	14	&	14	&	0.01 	&	&	14	&	0.0145	\\
\tabincell{c}{	p\_hat300-2	}&	33	&	41	&	0.06 	&	&	41	&	0.019	\\
\tabincell{c}{	p\_hat300-3	}&	43	&	59	&	0.09 	&	&	59	&	0.0545	\\
\tabincell{c}{	p\_hat500-1	}&	14	&	16	&	0.11 	&	&	16	&	0.0655	\\
\tabincell{c}{	p\_hat500-2	}&	-	&	57	&	0.07 	&	&	57	&	0.0255	\\
\tabincell{c}{	p\_hat500-3	}&	-	&	81	&	1.94 	&	&	81	&	0.136	\\
\tabincell{c}{	p\_hat700-1	}&	13	&	17	&	0.45 	&	&	17	&	0.2405	\\
\tabincell{c}{	p\_hat700-2	}&	50	&	70	&	0.26 	&	&	70	&	0.0735	\\
\tabincell{c}{	p\_hat700-3	}&	73	&	100	&	1.46 	&	&	100	&	0.144	\\
\tabincell{c}{	san1000	}&	-	&	33	&	1.70 	&	&	33(30.9)	&	0.0215	\\
\tabincell{c}{	san200\_0.7\_1	}&	-	&	60	&	0.01 	&	&	60	&	0.0155	\\
\tabincell{c}{	san200\_0.7\_2	}&	48	&	49	&	1.90 	&	&	49(48.95)	&	9.4395	\\
\tabincell{c}{	san200\_0.9\_1	}&	125	&	125	&	0.03 	&	&	125	&	0.022	\\
\tabincell{c}{	san200\_0.9\_2	}&	-	&	105	&	0.02 	&	&	105(104)	&	0.0245	\\
\tabincell{c}{	san200\_0.9\_3	}&	-	&	96	&	0.08 	&	&	96(91.5)	&	0.0245	\\
\tabincell{c}{	san400\_0.5\_1	}&	-	&	29	&	4.92 	&	&	29(26.25)	&	0.0075	\\
\tabincell{c}{	san400\_0.7\_1	}&	-	&	81(80.45)	&	17.54 	&	&	81(80.05)	&	0.0225	\\
\tabincell{c}{	san400\_0.7\_2	}&	-	&	61	&	0.57 	&	&	61(60.75)	&	36.4435	\\
\tabincell{c}{	san400\_0.7\_3	}&	-	&	50(49.45)	&	24.71 	&	&	50(47.4)	&	22.13	\\
\tabincell{c}{	san400\_0.9\_1	}&	-	&	200	&	0.12 	&	&	200(195)	&	0.05	\\
\tabincell{c}{	sanr200\_0.7	}&	-	&	30	&	0.14 	&	&	30	&	0.035	\\
\tabincell{c}{	sanr200\_0.9	}&	-	&	69	&	0.07 	&	&	69	&	0.0485	\\
\tabincell{c}{	sanr400\_0.5	}&	-	&	21	&	0.56 	&	&	21	&	0.1105	\\
\tabincell{c}{	sanr400\_0.7	}&	-	&	35	&	5.31 	&	&	35	&	5.346	\\
\hline
\end{tabular}
\end{scriptsize}
\end{center}
\end{table}
\renewcommand{\baselinestretch}{1.0}\large\normalsize

\renewcommand{\baselinestretch}{0.5}\large\normalsize
\begin{table}
\begin{center}
\begin{scriptsize}
\caption{Comparisons of BLS-RLE for the $k$-plex when $k$=5}
\label{tabledet5}
\begin{tabular}{lrrrrrr}
\hline
\tabincell{c}{Instance} &BKV   &\multicolumn{2}{c}{FD-TS} &&\multicolumn{2}{c}{BLS-RLE} \\
\cline{3-4} \cline{6-7}
 &   & Max(Avg.) & time(s)    & & Max(Avg.) & time(s)   \\
\hline
\tabincell{c}{	brock200\_1	}&	27	&	39	&	2.36 	&	&	39	&	0.966	\\
\tabincell{c}{	brock200\_2	}&	17	&	20	&	0.04 	&	&	20	&	0.022	\\
\tabincell{c}{	brock200\_3	}&	19	&	26	&	0.09 	&	&	26	&	0.0535	\\
\tabincell{c}{	brock200\_4	}&	21	&	30	&	0.20 	&	&	30	&	0.0565	\\
\tabincell{c}{	brock400\_1	}&	23	&	46(45.50)	&	36.57 	&	&	46	&	5.3855	\\
\tabincell{c}{	brock400\_2	}&	29	&	45	&	2.14 	&	&	45	&	0.245	\\
\tabincell{c}{	brock400\_3	}&	-	&	46(45.90)	&	55.95 	&	&	46	&	3.172	\\
\tabincell{c}{	brock400\_4	}&	30	&	46	&	25.67 	&	&	46	&	5.104	\\
\tabincell{c}{	brock800\_1	}&	-	&	37	&	27.12 	&	&	37	&	0.7105	\\
\tabincell{c}{	brock800\_2	}&	-	&	38(37.15)	&	32.46 	&	&	38(37.75)	&	52.655	\\
\tabincell{c}{	brock800\_3	}&	-	&	38(37.10)	&	36.78 	&	&	38	&	68.3545	\\
\tabincell{c}{	brock800\_4	}&	-	&	37	&	33.58 	&	&	37	&	4.0245	\\
\tabincell{c}{	C1000.9	}&	-	&	119(118.15)	&	60.39 	&	&	\textbf{122(121)}	&	53.142	\\
\tabincell{c}{	C125.9	}&	-	&	65	&	0.38 	&	&	65	&	0.02	\\
\tabincell{c}{	C2000.5	}&	-	&	28(27.15)	&	11.72 	&	&	28	&	33.969	\\
\tabincell{c}{	C2000.9	}&	-	&	132(129.65)	&	76.66 	&	&	\textbf{137(135.35)}	&	67.061	\\
\tabincell{c}{	C250.9	}&	-	&	84	&	4.72 	&	&	84	&	0.3695	\\
\tabincell{c}{	C4000.5	}&	-	&	29(28.20)	&	26.93 	&	&	\textbf{30(29.05)}	&	6.135	\\
\tabincell{c}{	C500.9	}&	-	&	103(102.25)	&	36.65 	&	&	\textbf{104(103.1)}	&	11.4765	\\
\tabincell{c}{	c-fat200-1	}&	14	&	14	&	0.00 	&	&	14	&	0.0145	\\
\tabincell{c}{	c-fat200-2	}&	24	&	24	&	0.01 	&	&	24(22.1)	&	0.0155	\\
\tabincell{c}{	c-fat200-5	}&	58	&	58	&	0.00 	&	&	58(56.95)	&	0.0175	\\
\tabincell{c}{	c-fat500-10	}&	126	&	126	&	0.16 	&	&	126(124.9)	&	0.026	\\
\tabincell{c}{	c-fat500-1	}&	15	&	15	&	0.00 	&	&	15(14.35)	&	0.0105	\\
\tabincell{c}{	c-fat500-2	}&	26	&	26	&	0.00 	&	&	26(25.3)	&	0.016	\\
\tabincell{c}{	c-fat500-5	}&	64	&	64	&	0.04 	&	&	64(62.7)	&	0.0145	\\
\tabincell{c}{	DSJC1000\_5	}&	-	&	27(26.25)	&	32.14 	&	&	27(26.85)	&	36.39	\\
\tabincell{c}{	DSJC500\_5	}&	-	&	24	&	1.07 	&	&	24	&	0.4795	\\
\tabincell{c}{	gen200\_p0.9\_44	}&	-	&	84	&	0.05 	&	&	84	&	0.0355	\\
\tabincell{c}{	gen200\_p0.9\_55	}&	-	&	80	&	0.19 	&	&	80	&	0.0815	\\
\tabincell{c}{	gen400\_p0.9\_55	}&	-	&	124	&	0.16 	&	&	124(122.65)	&	0.121	\\
\tabincell{c}{	gen400\_p0.9\_65	}&	-	&	138	&	0.15 	&	&	138(135)	&	0.0925	\\
\tabincell{c}{	gen400\_p0.9\_75	}&	-	&	136	&	0.12 	&	&	136(133.15)	&	0.0835	\\
\tabincell{c}{	hamming10-2	}&	512	&	513(512.15)	&	16.82 	&	&	513(512.1)	&	1.8635	\\
\tabincell{c}{	hamming10-4	}&	51	&	79(78.05)	&	34.37 	&	&	79(78.65)	&	41.8755	\\
\tabincell{c}{	hamming6-2	}&	48	&	48	&	0.00 	&	&	48	&	0.023	\\
\tabincell{c}{	hamming6-4	}&	12	&	12	&	0.00 	&	&	12	&	0.011	\\
\tabincell{c}{	hamming8-2	}&	128	&	152	&	0.97 	&	&	152	&	0.902	\\
\tabincell{c}{	hamming8-4	}&	20	&	32	&	0.02 	&	&	32	&	0.0185	\\
\tabincell{c}{	johnson16-2-4	}&	21	&	24	&	0.00 	&	&	24	&	0.0075	\\
\tabincell{c}{	johnson32-2-4	}&	-	&	48	&	0.09 	&	&	48	&	0.021	\\
\tabincell{c}{	johnson8-2-4	}&	12	&	12	&	0.00 	&	&	12	&	0.009	\\
\tabincell{c}{	johnson8-4-4	}&	24	&	28	&	0.00 	&	&	28	&	0.01	\\
\tabincell{c}{	keller4	}&	22	&	28	&	0.02 	&	&	28	&	0.0505	\\
\tabincell{c}{	keller5	}&	-	&	61	&	5.04 	&	&	61(60.55)	&	10.423	\\
\tabincell{c}{	keller6	}&	-	&	125(123.20)	&	73.91 	&	&	125(124.05)	&	58.103	\\
\tabincell{c}{	MANN\_a27	}&	351	&	351	&	0.45 	&	&	351	&	0.0675	\\
\tabincell{c}{	MANN\_a45	}&	990	&	990	&	7.44 	&	&	990	&	2.507	\\
\tabincell{c}{	MANN\_a81	}&	-	&	3135(2660.75)	&	190.01 	&	&	\textbf{3240} 	&	73.26 	\\
\tabincell{c}{	MANN\_a9	}&	44	&	45	&	0.00 	&	&	45	&	0.0005	\\
\tabincell{c}{	p\_hat1000-1	}&	-	&	20	&	7.17 	&	&	20(19.5)	&	0.5625	\\
\tabincell{c}{	p\_hat1000-2	}&	-	&	84	&	1.30 	&	&	84	&	0.13	\\
\tabincell{c}{	p\_hat1000-3	}&	-	&	122	&	29.84 	&	&	122	&	0.536	\\
\tabincell{c}{	p\_hat1500-1	}&	-	&	21	&	1.43 	&	&	21	&	0.6005	\\
\tabincell{c}{	p\_hat1500-2	}&	-	&	117(116.55)	&	41.90 	&	&	117	&	3.613	\\
\tabincell{c}{	p\_hat1500-3	}&	-	&	164	&	48.76 	&	&	\textbf{165(164.9)}	&	47.2155	\\
\tabincell{c}{	p\_hat300-1	}&	14	&	16	&	0.02 	&	&	16	&	0.0195	\\
\tabincell{c}{	p\_hat300-2	}&	33	&	46	&	0.03 	&	&	46	&	0.0205	\\
\tabincell{c}{	p\_hat300-3	}&	43	&	65	&	0.11 	&	&	65	&	0.0285	\\
\tabincell{c}{	p\_hat500-1	}&	14	&	18	&	0.15 	&	&	18	&	0.076	\\
\tabincell{c}{	p\_hat500-2	}&	-	&	62	&	0.25 	&	&	62	&	0.049	\\
\tabincell{c}{	p\_hat500-3	}&	-	&	89	&	1.73 	&	&	\textbf{90(89.95)}	&	35.301	\\
\tabincell{c}{	p\_hat700-1	}&	13	&	19	&	2.00 	&	&	19	&	14.0355	\\
\tabincell{c}{	p\_hat700-2	}&	50	&	79	&	9.42 	&	&	79	&	0.7935	\\
\tabincell{c}{	p\_hat700-3	}&	73	&	109	&	1.35 	&	&	109	&	0.112	\\
\tabincell{c}{	san1000	}&	-	&	41	&	6.39 	&	&	41(39.15)	&	0.0415	\\
\tabincell{c}{	san200\_0.7\_1	}&	-	&	75	&	0.01 	&	&	75	&	0.0145	\\
\tabincell{c}{	san200\_0.7\_2	}&	48	&	60	&	0.02 	&	&	60	&	0.011	\\
\tabincell{c}{	san200\_0.9\_1	}&	125	&	125	&	0.03 	&	&	125	&	0.0275	\\
\tabincell{c}{	san200\_0.9\_2	}&	-	&	105	&	0.03 	&	&	105(104.5)	&	0.0255	\\
\tabincell{c}{	san200\_0.9\_3	}&	-	&	100	&	0.03 	&	&	100	&	0.019	\\
\tabincell{c}{	san400\_0.5\_1	}&	-	&	36(35.70)	&	55.40 	&	&	\emph{35(33.75)}	&	0.016	\\
\tabincell{c}{	san400\_0.7\_1	}&	-	&	100	&	0.06 	&	&	100	&	0.0325	\\
\tabincell{c}{	san400\_0.7\_2	}&	-	&	76	&	10.34 	&	&	76(75.25)	&	5.833	\\
\tabincell{c}{	san400\_0.7\_3	}&	-	&	61	&	0.29 	&	&	61(57.65)	&	14.6585	\\
\tabincell{c}{	san400\_0.9\_1	}&	-	&	200	&	0.15 	&	&	200	&	0.059	\\
\tabincell{c}{	sanr200\_0.7	}&	-	&	33	&	0.05 	&	&	33	&	0.028	\\
\tabincell{c}{	sanr200\_0.9	}&	-	&	77	&	5.74 	&	&	77	&	0.408	\\
\tabincell{c}{	sanr400\_0.5	}&	-	&	24	&	1.63 	&	&	24	&	0.1875	\\
\tabincell{c}{	sanr400\_0.7	}&	-	&	39	&	31.01 	&	&	39	&	2.3455	\\
\hline
\end{tabular}
\end{scriptsize}
\end{center}
\end{table}
\renewcommand{\baselinestretch}{1.0}\large\normalsize

\end{document}